\documentclass[a4paper]{article}

\usepackage{INTERSPEECH_v2}

\usepackage{graphicx}
\usepackage{amssymb,amsmath,bm}
\usepackage{textcomp}
\usepackage[inline]{enumitem}
\usepackage{graphics}
\usepackage{multirow}
\usepackage{hyperref}

\def\vec#1{\ensuremath{\bm{{#1}}}}

\ninept

\title{Learning Latent Representations for Speech Generation and Transformation}

\name{Wei-Ning Hsu, Yu Zhang, James Glass}
\address{Computer Science and Artificial Intelligence Laboratory \\
         Massachusetts Institute of Technology\\
         Cambridge, MA 02139, USA}
\email{\{wnhsu,yzhang87,jrg\}@csail.mit.edu}

\begin{document}

  \maketitle
  
  \begin{abstract}
  An ability to model a generative process and learn a latent representation for speech in an unsupervised fashion will be crucial to process vast quantities of unlabelled speech data. Recently, deep probabilistic generative models such as Variational Autoencoders (VAEs) have achieved tremendous success in modeling natural images. In this paper, we apply a convolutional VAE to model the generative process of natural speech. We derive latent space arithmetic operations to disentangle learned latent representations. We demonstrate the capability of our model to modify the phonetic content or the speaker identity for speech segments using the derived operations, without the need for parallel supervisory data.
  \end{abstract}
  

  \section{Introduction}
  Speech waveforms have complex distributions that exhibit high variance due to factors that include linguistic content, speaking style, dialect, speaker identity, emotional state, environment, channel effects, etc. Understanding the influence of these factors on the speech signal is an important problem, which can be used for a wide variety of applications, including, but not limited to adaptation and data augmentation for speech recognition \cite{Jaitly_vocaltract13,Cui15}, voice conversion \cite{Kain98spectralvoice,Stylianou09,Toda06}, and speech compression \cite{wong1983very}. However, most previous research has focused on handcrafting features to capture these factors, rather than learning these factors automatically through a probabilistic generative process.
  
  
  Recently, there has been significant interest in deep probabilistic generative models, such as Variational Autoencoders (VAEs)~\cite{kingma2013auto} and Generative Adversarial Nets (GANs)~\cite{goodfellow2014generative}. Particularly, VAE addresses the intractability issue that occurs in even moderately complicated models such as Restricted Boltzmann machines (RBMs), which have been applied for voice conversion~\cite{Zhizheng13,Nakashika2015,Nakashika2016}, and provides efficient approximated posterior inference of the latent factors.
  While there are many works investigating generative models for natural images~\cite{radford2015unsupervised, larsen2015autoencoding}, little work has been done on learning speech generation with deep probabilistic generative models~\cite{blaauw2016modeling,hsu2016voice}.

  In this paper, we adopt the VAE framework and propose a convolutional architecture to model the probabilistic generative process of speech to learn a latent representation. We present simple arithmetic operations in the latent space to demonstrate that such operations can decompose the latent representation into different attributes, such as speaker identity and linguistic content. By manipulating the latent representation, we also demonstrate an ability to perturb some aspect of the surface speech segment, for example the speaker identity, while keeping the remaining attributes fixed (e.g., linguistic content). To quantify the behavior of the latent representation modifications, an experiment is conducted to measure our ability to modify speaker characteristics without changing linguistic content, and vice versa.  In addition, we perform an analysis to evaluate the model's ability to generate speech segments of different durations.
  
  The rest of the paper is organized as follows. In Section~\ref{sec:relate}, we briefly discuss related work. Our models and an analysis of latent representations are detailed in Section~\ref{sec:model} and \ref{sec:analysis}.
  Data preparation is explained in Section~\ref{sec:data}. In Section~\ref{sec:exp}, we show the experimental results. Finally, we conclude our work and discuss our future research plans on this topic in Section \ref{sec:conclusion}.

  \section{Related Work}\label{sec:relate}
  Recent research on speech and audio generation has made remarkable progress on directly utilizing time-domain speech signals. WaveNet~\cite{van2016wavenet} introduces the one-dimensional dilated causal convolutional model, where the effective receptive field grows exponentially wide with the depth by using exponentially growing dilation factors with the depth.  A different model called SampleRNN~\cite{mehri2016samplernn} presented a multi-scale recurrent neural network, where each layer is operated at different clock rates and each sample is generated conditioned on all the previous samples. Both models focused on generating high quality audio segments by predicting the next sample given the preceding samples, instead of learning latent representations for the entire audio segments using probabilistic generative models.
  
  While VAEs have been widely applied for image generation, there has been less speech research on this topic. A VAE-based framework was used in \cite{tan2016learning} to extract both frame-level and utterance-level features that were used in combination with other features for robust speech recognition.  A fully-connected VAE was used in \cite{blaauw2016modeling} to learn a frame-level latent representation, and evaluated using a Gaussian diffusion process to generate and concatenate multiple samples that varied smoothly in time.
  
  
  

  
  \section{Model}\label{sec:model}
  \subsection{Variational Autoencoder} \label{sec:vae}
  
  Variational autoencoders \cite{kingma2013auto} define a probabilistic generative process between observation $\bm{x}$ and latent variable $\bm{z}$ as follows: $\bm{z} \sim p_{\bm{\theta}^*}(\bm{z})$ and $\bm{x} \sim p_{\bm{\theta}^*}(\bm{x}|\bm{z})$,
  where the prior $p_{\bm{\theta}^*}(\bm{z})$ and the conditional likelihood $p_{\bm{\theta}^*}(\bm{x}|\bm{z})$ are from a probability distribution family parameterized by $\bm{\theta}$. In an unsupervised setting, we are only given a dataset $\bm{X} = \{ \bm{x}^{(i)} \}_{i=1}^N$, so the true value of $\bm{\theta^*}$, as well as the latent variable $\bm{z}$ for each observation $\bm{x}$ in this process are unknown. 
  
  We are often interested in knowing the marginal likelihood of the data $p_{\bm{\theta}}(\bm{x})$, or the posterior  $p_{\bm{\theta}}(\bm{z}|\bm{x})$; however, both require computing the intractable integral $\int p_{\bm{\theta}}(\bm{z}) p_{\bm{\theta}}(\bm{x}|\bm{z}) d\bm{z}$. To solve this problem, VAEs introduce a recognition model $q_{\bm{\phi}}(\bm{z}|\bm{x})$, which approximates the true posterior $p_{\bm{\theta}}(\bm{z}|\bm{x})$. We can therefore rewrite the marginal likelihood as:
  \begin{align}
      &\log p_{\bm{\theta}}(\bm{x}) = D_{KL}(q_{\bm{\phi}}(\bm{z}|\bm{x})||p_{\bm{\theta}}(\bm{z}|\bm{x})) + \mathcal{L}(\bm{\theta},\bm{\phi};\bm{x}) \nonumber\\
      &\ge \mathcal{L}(\bm{\theta},\bm{\phi};\bm{x}) \nonumber\\
      &= - D_{KL}(q_{\bm{\phi}}(\bm{z}|\bm{x}) || p_{\bm{\theta}}(\bm{z}) ) +  \mathbb{E}_{q_{\bm{\phi}}(\bm{z}|\bm{x})} [ \log p_{\bm{\theta}}(\bm{x}|\bm{z}) ] \label{eq:lb},
  \end{align}
  where $\mathcal{L}(\bm{\theta}, \bm{\phi}; \bm{x})$ is the variational lower bound we want to optimize with respect to $\bm{\theta}$ and $\bm{\phi}$.
  
  In the VAE framework we consider here, both the recognition model $q_{\bm{\phi}}(\bm{z}|\bm{x})$ and the generative model $p_{\bm{\theta}}(\bm{x}|\bm{z})$ are parameterized using diagonal Gaussian distributions, of which the mean and the covariance are computed with a neural network. The prior is assumed to be a centered isotropic multivariate Gaussian $p_{\bm{\theta}}(\bm{z}) = \mathcal{N}(\bm{z}; \bm{0}, \bm{I})$, that has no free parameters. 
  
  In practice, the expectation in (\ref{eq:lb}) is approximated by first drawing $L$ samples from $\bm{z}^{l} \sim q_{\bm{\phi}}(\bm{z}|\bm{x})$, and then computing $\mathbb{E}_{q_{\bm{\phi}}(\bm{z}|\bm{x})} [ \log p_{\bm{\theta}}(\bm{x}|\bm{z}) ] \simeq \frac{1}{L} \sum_{l=1}^{L} \log p_{\bm{\theta}}(\bm{x}|\bm{z}^l)$. To yield a differentiable network after sampling, the reparameterization trick~\cite{kingma2013auto} is used. Suppose $\bm{z} \sim \mathcal{N}(\bm{z}; \bm{\mu_z}, \bm{\sigma_z}^2\bm{I})$, after reparameterizing we have $\bm{z} = \bm{\mu_z} + \bm{\sigma_z} \odot \bm{\epsilon}$, where $\odot$ denotes an element-wise product, and vector $\bm{\epsilon}$ is sampled from $\mathcal{N}(\bm{0}, \bm{I})$ and treated as an additional input.
  
  \subsection{Proposed Model Architecture}\label{sec:arch}
  In this work, our goal is to learn latent representations of speech segments to model the generation process. We let the observed data $\bm{x}$ be a sequence of frames of fixed length. The learned latent variable $\bm{z}$ is therefore supposed to encode the factors that result in the variability of speech segments, such as the content being spoken, speaker identity, and channel effect. 
  
  As mentioned earlier, a VAE is composed of two networks: a recognition network, and a generative network. The recognition network takes a speech segment as input and predicts the mean $\bm{\mu_z}$ and the log-variance $\log\bm{\sigma_z}^2$ that parameterize the posterior distribution $q_{\bm{\phi}}(\bm{z}|\bm{x})$. A speech segment is treated as a two dimensional image of width $T$ and height $F$; however, unlike natural images, speech segments are only translational invariant to the time axis. Therefore, similar to \cite{harwath2017learning}, 1-by-$F$ filters are applied at the first convolutional layer, and $w$-by-1 filters at following layers. As suggested in \cite{radford2015unsupervised}, instead of pooling, we use stride size $>1$ for down-sampling along the time axis. The output from the last convolutional layer is flattened and fed into fully connected layers before going to the Gaussian parameter layer modeling the latent variable $\bm{z}$. See Table \ref{tab:model} for a summary.
  
  The generative network takes sampled $\bm{z}$ as input, and predicts the mean $\bm{\mu_x}$ as well as the log-variance $\log\bm{\sigma_x}^2$ of the observed data. Here we use symmetric architectures to the corresponding recognition network. 
  
  \begin{table}[tbh]
      \centering
      \begin{tabular}{c|c|c|c|c|c}
           & Conv1 & Conv2 & Conv3 & Fc1 & Gauss \\
           \hline
           \#filters/units  & 64 & 128 & 256 & 512 & 128\\
           filter size      & 1x$F$ & 3x1 & 3x1 & - & -\\
           stride           & (1,1) & (2,1) & (2,1) & - & -\\
      \end{tabular}
      \caption{Recognition network architecture. \textit{Conv} refers to convolutional layers, \textit{Fc} refers to fully connected layers, and \textit{Gauss} refers to the Gaussian parametric layer modeling $\bm{z}$}
      \label{tab:model}
  \end{table} 
  
  Different choices for the activation function were investigated. No activation is applied to Gaussian parameter layers, since the mean and the log-variance are unbounded for both $\bm{x}$ and $\bm{z}$. For other layers, we use $\tanh$ because the unbounded rectifier linear units led to overflow of the KL-divergence and conditional likelihood. Batch normalization is applied to every layer except for the Gaussian parameter layer.
  
  
  
    
  \section{Latent Representation Analysis}\label{sec:analysis}

    In this section, we examine how to decompose speech attributes from the learned latent representations. Here, we use $a$ to denote the attribute and $r$ to denote the value of the attribute.
    
  \subsection{Deriving Latent Attribute Representations}\label{sec:lar}
    The first assumption we make is that conditioning on some attribute $a$ being $r$, such as the \textit{phone} being \textit{/ae/}, the prior distribution of $\bm{z}$ is also a Gaussian; in other words, $p(\bm{z};a=r) = N(\bm{z}; \bm{\mu}_r, \bm{\Sigma}_r)$. We therefore define $\bm{\mu}_r$ as the \textit{latent attribute representation} for $r$.
    Let $\bm{X}_r=\{\bm{x}_r^{(i)}\}_{i=1}^{N_r}$ be a subset of $\bm{X}$ where the attribute $a$ of each instance is $r$. We can then estimate $\bm{\mu}_r$ as follows:
      \begin{align}
        \bm{\mu}_r &= \mathbb{E}_{p_{\theta}(\bm{z}; r)}[\bm{z}] = \int_{\bm{z}}\int_{\bm{x}} \bm{z} p_{\theta}(\bm{z}|\bm{x};r) p_{\theta}(\bm{x};r) \nonumber\\
        &\approx \int_{\bm{z}}\int_{\bm{x}} \bm{z} q_{\bm{\phi}}(\bm{z}|\bm{x};r) p_{\bm{\theta}}(\bm{x};r) \approx \frac{1}{N_r} \sum_{i=1}^{N_r} \int_z \bm{z} q_{\phi}(\bm{z}|\bm{x}_r^{(i)}) \nonumber\\
            &\approx \dfrac{1}{N_r} \sum_{i=1}^{N_r} (\dfrac{1}{J} \sum_{j=1}^{J}\bm{z}^{(i,j)}),
      \end{align}
    where $\bm{z}^{(i,j)} \sim q_{\bm{\phi}}(\bm{z}|\bm{x}_r^{(i)})$. This results in averaging the $J$ sampled latent representations of each instance in $\bm{X}_r$.
    Furthermore, let $\bar{\bm{z}} = \frac{1}{N}\sum_{i=1}^N \bm{z}_i$, since $p(\bm{z})$ is a log-concave function, it is guaranteed that $p(\bar{\bm{z}}) > \min_{\bm{z}_i}p(\bm{z}_i)$. Therefore VAE should be able to generate reasonable speech-like segment from $\bar{\bm{z}}$ if all the $p(\bm{z}_i)$ have high values. 
    
    \subsection{Arithmetic Operations to Modify Speech Attributes}\label{sec:mod}
    Here we make the second assumption: let there be $K$ independent attributes that affect the realization of speech, each attribute $a_k$ is then modeled using a subspace $\mathit{Z}_{a_k}$, where $\mathit{Z} = \cup_{k=1}^K \mathit{Z_{a_k}}$ and $\mathit{Z}_{a_k} \perp \mathit{Z}_{a_{k'}}$ if $k \neq k'$. 
    Hence, the latent representation can be decomposed into $K$ orthogonal latent attribute representations $\bm{z}_{a_1}, \bm{z}_{a_2}, \cdots, \bm{z}_{a_k}$, where $\bm{z}_{a_k} \in \mathit{Z}_{a_k}$ and $\bm{z} = \sum_{k=1}^{K}\bm{z}_{a_k}$. 
    Combining the aforementioned assumption of the conditioned prior of $\bm{z}$, we can next derive the latent space arithmetic operations to modify the speech attributes. 
        
    Suppose we want to modify the attribute $a_k$, for example the speaker identity, of a speech segment $\bm{x}^{(i)}$, from being speaker $r_s$ to being speaker $r_t$. Given the latent attribute representations $\bm{\mu}_{r_s}$ and $\bm{\mu}_{r_t}$ for speaker $r_s$ and $r_t$ respectively, the \textit{latent attribute shift} $\bm{v}_{r_s \rightarrow r_t}$ is computed as: $\bm{v}_{r_s \rightarrow r_t} = \bm{\mu}_{r_t} - \bm{\mu}_{r_s}$. We can then modify the speech $\bm{x}^{(i)}$ as follows:
    \begin{align}
      \bm{z}^{(i)} &\sim q_{\bm{\phi}}(\bm{z}|\bm{x}^{(i)}) \\
      \bm{z}_{mod}^{(i)} &= \bm{z}^{(i)} + \bm{v}_{r_s \rightarrow r_t} \\
      \bm{x}_{mod}^{(i)} &\sim p_{\bm{\theta}}(\bm{x}|\bm{z}_{mod}^{(i)}),
    \end{align}
    which does not modify latent attribute representations other than $\bm{z}_{a_k}^{(i)}$, because $\bm{v}_{r_s \rightarrow r_t} \perp \bm{z}_{a_{k'}}^{(i)}$ for $k' \neq k$.

  \section{Data}\label{sec:data}
  
  \subsection{TIMIT}
  The TIMIT acoustic-phonetic corpus~\cite{garofolo1993darpa,zue1990speech} contains broadband recordings of phonetically-balanced read speech. A total of 6300 utterances (5.4 hours) are presented with 10 sentences from each of 630 speakers, of which approximately 70\% are male and 30\% are female. Each utterance comes with manually time-aligned phonetic and word transcriptions, as well as a 16-bit, 16kHz speech waveform file. We follow Kaldi's TIMIT recipe to split train/dev/test sets and exclude dialect sentences (SA), with 462/50/24 non-overlapping speakers in each set respectively. Phonetic transcriptions are based on 58 phones, excluding silence phones. 

  \subsection{Data Preprocessing}
  We consider two types of frame representations: magnitude spectrum in dB (Spec) and filter banks (FBank). For both features, we first apply a 25ms Hanning window with 10ms shift, and then compute the short time Fourier transform coefficients with flooring at -20dB. 
  For FBank features, 80 Mel-scale filter banks that match human perceptual sensitivity are applied, which preserves more detail at lower frequency regions.
  
  
  We investigate two different segment lengths: 200ms and 1s, which correspond to 20 frames and 100 frames, and are referred to as \textit{syllable-level} and \textit{word-level} datasets, respectively. 
  
  

  \section{Experimental Results}\label{sec:exp}
  \subsection{Experiment Setups}
  All models were trained with stochastic gradient descent using a mini-batch size of 128 without clipping to minimize the negative variational lower bound plus an  $L2$-regularization with weight $10^{-4}$. The Adam~\cite{kingma2014adam} optimizer is used with $\beta_1=0.95$, $\beta_2=0.999$, $\epsilon=10^{-8}$, and initial learning rate of $10^{-3}$. Training is terminated if the lower bound on the development set does not improve for 10 epochs. To compare with VAE, we also train an autoencoder (AE) with the same proposed model architecture except for the Gaussian latent variable layer, which is replaced with a fully-connected layer of 128 hidden units\footnote{Both VAE and AE models show reasonable reconstruction performance on both Fbank and Spec. We do not show the reconstructed features in this section due to space limitations.}.

  \subsection{Latent Attribute Representation}
    
\begin{figure}[tbh]
    \centering
    \begin{minipage}[b]{.32\linewidth}
      \centering
      \centerline{\includegraphics[width=1.0\linewidth]{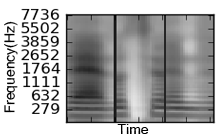}}
      \centerline{(a) VAE}
    \end{minipage}
    \begin{minipage}[b]{.32\linewidth}
      \centering
      \centerline{\includegraphics[width=1.0\linewidth]{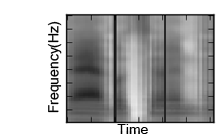}}
      \centerline{(b) AE}
    \end{minipage}
    \begin{minipage}[b]{.32\linewidth}
      \centering
      \centerline{\includegraphics[width=1.0\linewidth]{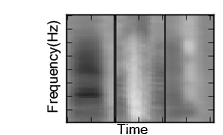}}
      \centerline{(c) Fbank}
    \end{minipage}
    \caption{Comparison between VAE, AE and Fbank on averaging representations of /ae/, /th/, and /n/ from left to right. Each segment is 200ms long.}
    \label{fig:avg}
    \end{figure}

    In Figure \ref{fig:avg}, we show the results of reconstructing from latent attribute representations of three phones, /ae/, /th/, and /n/, using VAE and AE respectively, based on the derivation in Section \ref{sec:lar}. As a baseline, we also show the results of averaging filter bank features. The VAE preserves more harmonic structure and clearer spectral envelope, while the AE and the Fbank are more blurred. It is worth noting that AE also shows unnatural frequent vertical stripe artifacts.
    
    \subsection{Modifying Attributes of Speech}
      
\begin{figure}[!h]
      \centering
      \includegraphics[width=0.75\linewidth]{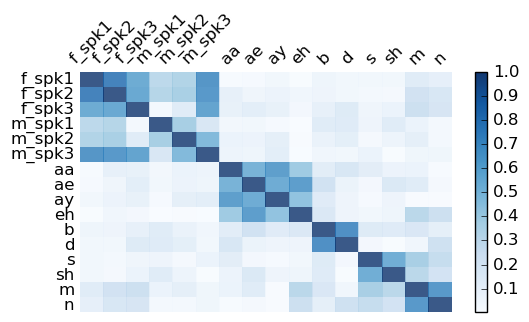}
      \caption{Cosine similarities of latent attribute representations.}
      \label{fig:cos_sim}
    \end{figure}

    To assess the orthogonality-between-attributes assumption, we sampled six speakers, three males and three females, denoted by \textit{\{m,f\}\_spk[i]}, and ten phones, including vowels, stops, fricatives, and nasals, to compute three latent speaker representations and ten latent phone representations. Figure~\ref{fig:cos_sim} plots the cosine similarities between these representations. 
    From the figure, we can observe that off-diagonal blocks have low cosine similarities, which indicates that latent speaker representations and latent phone representations reside in orthogonal latent subspaces. Second, different latent phone representations also cluster according to the phonetic characteristics, which suggests the latent phone subspace may be further divided.
        
\begin{figure}[!ht]
      \centering
      \includegraphics[width=.95\linewidth]{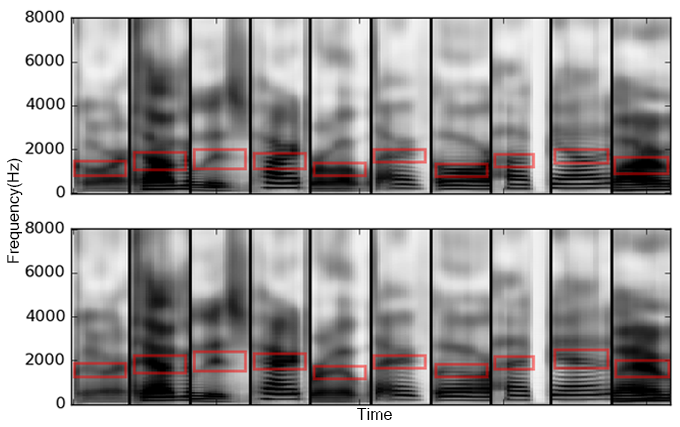}
      \caption{Modify the phone from /aa/ (top) to /ae/ (bottom). Each segment is 200ms long.}
      \label{fig:mod_phone}
    \end{figure}

    We next explored modifying the phone and speaker attributes using the derived operations in Section~\ref{sec:mod}.\footnote{More sound examples can be found at: \url{http://people.csail.mit.edu/wnhsu/vae_speech}} 
    Figure \ref{fig:mod_phone} shows an example of drawing 10 instances of the phone /aa/ and transforming them to /ae/ using the latent attribute shift $\bm{v}_{aa \rightarrow ae}$. We can clearly observe that the second formant $F_2$, marked with red boxes,\footnote{Best viewed in color} of each instance goes up after modification, because it is being changed from a back vowel to a front vowel. On the other hand, the harmonics of each instance, which are closely related to the speaker identity, maintain roughly the same. 
    
\begin{figure}[!h]
      \centering
      \includegraphics[width=.95\linewidth]{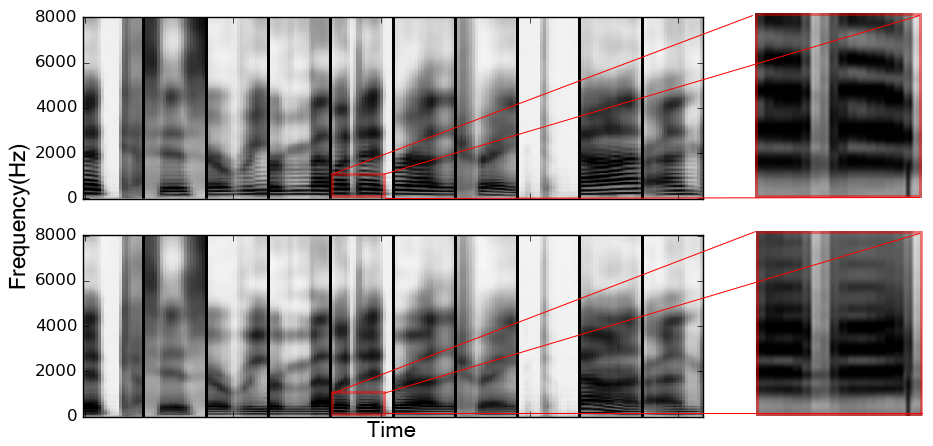}
      \caption{Modify from a female (top) to a male (bottom). Each segment is 200ms long.}
      \label{fig:mod_spk}
    \end{figure}

    Figure \ref{fig:mod_spk} illustrates modifying 10 instances from a female speaker \textit{falk0} to a male speaker \textit{madc0} with the latent attribute shift $\bm{v}_{falk0 \rightarrow madc0}$. The harmonics (horizontal stripes) decrease after modification, while the spectrum envelope remains the same, indicating that the phonetic content is not changed.

    In an attempt to quantify our latent attribute perturbation, we trained convolutional phonetic and speaker classifiers so that we could measure the difference of the posterior of each attribute before and after modification. The 58-class phone classifier achieves a test accuracy of $72.2\%$, while the 462-class speaker classifier achieves a test accuracy of $44.2\%$.
    
    The shifts in posterior distributions of the phone and speaker classifications on the modified data are shown in Table \ref{table:mod}.  The upper half of the table contains results for speech segments that were transformed from /aa/ to /ae/.  The first row shows that the average /aa/ posterior was $34\%$ while the average correct speaker posterior was $51\%$.  The second row shows that after modification to an /ae/, the average phone posteriors shift dramatically to be $30\%$ /ae/, while slightly degrading the average correct speaker posterior.
    
    The lower part of the table shows the results of speech segments that had speaker identity modified from speaker `falk0' to `madc0'.  The third row shows an average speaker posterior of $44\%$ for `falk0' in the unmodified samples, while the average correct phone posterior was $55\%$.  After modification we see that the average speaker posterior has shifted to be $29\%$ `madc0' while slightly degrading the average correct phone posterior.
    \begin{table}[h]
    \centering
    \begin{tabular}{c|c|c|c||c}
      \hline
      \multirow{3}{*}{Modify Phone}&& /aa/ & /ae/ & ori. spk. \\      
      \cline{2-5} & before   & 34.06\%   & 0.45\%    & 50.78\% \\
                  & after    & 0.24\%    & 29.73\%   & 41.66\% \\
      \hline
      \hline
      \multirow{3 }{*}{Modify Speaker}&& falk0 & madc0 & ori. phone \\
       
      \cline{2-5} & before    & 44.48\%   & 0.02\%    & 54.61\% \\
                  & after     & 3.11\%    & 28.71\%   & 48.71\% \\
      \hline
    \end{tabular}
    \caption{Average posteriors over 10 instances of source, target, and fixed attributes before and after modification.}
    \label{table:mod}
    \end{table}
        
    \subsection{Random Sampling from the Latent Space}
    One of the advantages of VAEs is that the prior $p_\theta(z)$ is assumed to be a centered isotropic Gaussian, which enables us to sample latent vectors and reconstruct speech-like segments. Here, we investigate the syllable-level and word-level datasets.
    
    \begin{figure}[h]
    \centering
    \begin{minipage}[b]{.48\linewidth}
      \centering
      \centerline{\includegraphics[width=1.0\linewidth]{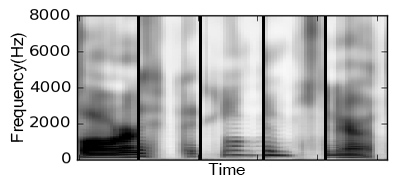}}
      \centerline{(a) Syllable-level}
    \end{minipage}
    \begin{minipage}[b]{.48\linewidth}
      \centering
      \centerline{\includegraphics[width=1.0\linewidth]{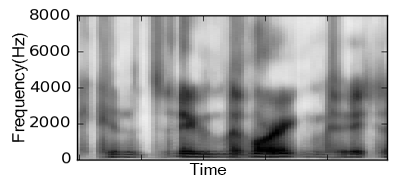}}
      \centerline{(b) Word-level}
    \end{minipage}
    \caption{Random samples drawn from models trained with syllable-level and word-level dataset. The segments in (a) are 200ms, and the segment in (b) is 1s.}
    \label{fig:rand}
    \end{figure}
\begin{figure}[]
      \centering
      \includegraphics[width=\linewidth]{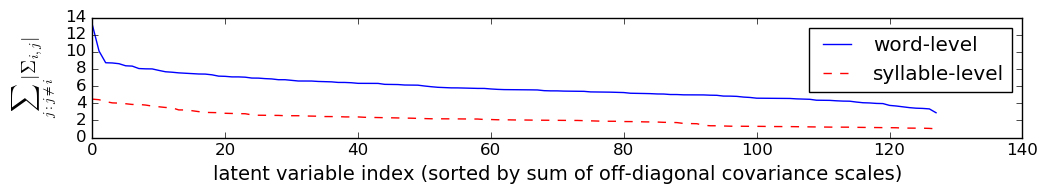}
      \caption{Comparison of sum of off-diagonal covariance scales for each dimension for the syllable and word-level dataset.}
      \label{fig:cov}
    \end{figure}

    Figure \ref{fig:rand} (a) shows five random samples from the syllable-level model, which look and sound reasonable; however, we observe that random samples drawn from the word-level model are less natural because of excessive closures (vertical stripes), as shown in Figure \ref{fig:rand} (b). The failure from drawing random samples implies that there is discrepancy between the assumed prior and the true prior. We hypothesize that because per-dimension KL-divergence values are computed, and correlations among dimensions are not penalized, the covariance matrix of the true prior may not be diagonal. We estimate the covariance matrix of the true prior by sampling the latent representations of the entire test set and compute the full covariance matrix. Figure \ref{fig:cov} compares the syllable model and the word model on the sum of off-diagonal covariance scale for each dimension. We can observe that the word-level model has higher correlations between different dimensions than the syllable-level model.
    
    \subsection{Walking in the Latent Space}
\begin{figure}[tbh]
        \centering
        \begin{minipage}[b]{.46\linewidth}
          \centering
          \centerline{\includegraphics[width=1.0\linewidth]{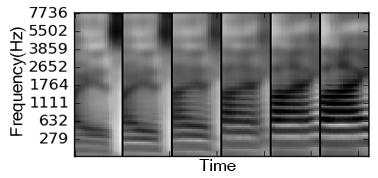}}
          \centerline{(a) VAE}
        \end{minipage}
        \begin{minipage}[b]{.46\linewidth}
          \centering
          \centerline{\includegraphics[width=1.0\linewidth]{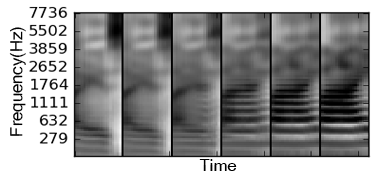}}
          \centerline{(b) AE}
        \end{minipage}
        \caption{Interpolation in the latent space using VAE and AE. Each segment is 200ms long.}
        \label{fig:trans}
    \end{figure}

    Finally, we explore the operation of interpolation in the latent space between speech segments. Since $p(\bm{z})$ is log-concave, the interpolated $\bm{z}_{int} = \alpha\bm{z}_a + (1-\alpha)\bm{z}_b$, where $\alpha \in [0, 1]$, would have $p(\bm{z}_{int}) \geq \min(p(\bm{z}_a), p(\bm{z}_b))$. Therefore it should also generate reasonable speech-like segments. Figure \ref{fig:trans} shows the transition between a male /ey/ to a female /ay/ using VAE and AE respectively. For VAE, we can clearly observe the pitch shifting and the formant contour transforming; however for AE it is more akin to interpolation in the raw feature space, where the magnitude of one segment goes down as the other goes up.
    
  \section{Conclusions and Future Work}
  \label{sec:conclusion}
  In this paper, we present a convolutional VAE to model the speech generation process, and learn latent representations for speech in an unsupervised framework. The abilities to decompose the learned latent representations and modify attributes of speech segments are demonstrated qualitatively and quantitatively. For future work, we plan to extend to hierarchical recurrent models in order to capture information at different time scales, and generate speech of variable lengths. 
  

  \newpage
  \eightpt
  \bibliographystyle{IEEEtran}

  \bibliography{main.bib}

\begin{thebibliography}{10}
\providecommand{\url}[1]{#1}
\csname url@samestyle\endcsname
\providecommand{\newblock}{\relax}
\providecommand{\bibinfo}[2]{#2}
\providecommand{\BIBentrySTDinterwordspacing}{\spaceskip=0pt\relax}
\providecommand{\BIBentryALTinterwordstretchfactor}{4}
\providecommand{\BIBentryALTinterwordspacing}{\spaceskip=\fontdimen2\font plus
\BIBentryALTinterwordstretchfactor\fontdimen3\font minus
  \fontdimen4\font\relax}
\providecommand{\BIBforeignlanguage}[2]{{%
\expandafter\ifx\csname l@#1\endcsname\relax
\typeout{** WARNING: IEEEtran.bst: No hyphenation pattern has been}%
\typeout{** loaded for the language `#1'. Using the pattern for}%
\typeout{** the default language instead.}%
\else
\language=\csname l@#1\endcsname
\fi
#2}}
\providecommand{\BIBdecl}{\relax}
\BIBdecl

\bibitem{Jaitly_vocaltract13}
N.~Jaitly and G.~E. Hinton, ``Vocal tract length perturbation ({VTLP}) improves
  speech recognition,'' in \emph{ICML workshop on Deep Learning for Audio,
  Speech, and Language Processing}, 2013.

\bibitem{Cui15}
X.~Cui, V.~Goel, and B.~Kingsbury, ``Data augmentation for deep neural network
  acoustic modeling,'' \emph{IEEE/ACM Trans. Audio, Speech and Lang. Proc.},
  vol.~23, no.~9, pp. 1469--1477, 2015.

\bibitem{Kain98spectralvoice}
A.~Kain and M.~W. Macon, ``Spectral voice conversion for text-to-speech
  synthesis,'' in \emph{ICASSP}, 1998.

\bibitem{Stylianou09}
Y.~Stylianou, ``Voice transformation: A survey.'' in \emph{ICASSP}.\hskip 1em
  plus 0.5em minus 0.4em\relax IEEE, 2009, pp. 3585--3588.

\bibitem{Toda06}
T.~Toda, Y.~Ohtani, and K.~Shikan, ``Eigenvoice conversion based on gaussian
  mixture model,'' in \emph{Interspeech}, 2006, pp. 2446--2449.

\bibitem{wong1983very}
D.~Wong, B.~Juang, and D.~Cheng, ``Very low data rate speech compression with
  {LPC} vector and matrix quantization,'' in \emph{Acoustics, Speech, and
  Signal Processing, IEEE International Conference on ICASSP'83.},
  vol.~8.\hskip 1em plus 0.5em minus 0.4em\relax IEEE, 1983, pp. 65--68.

\bibitem{kingma2013auto}
D.~P. Kingma and M.~Welling, ``Auto-encoding variational bayes,'' \emph{arXiv
  preprint arXiv:1312.6114}, 2013.

\bibitem{goodfellow2014generative}
I.~Goodfellow, J.~Pouget-Abadie, M.~Mirza, B.~Xu, D.~Warde-Farley, S.~Ozair,
  A.~Courville, and Y.~Bengio, ``Generative adversarial nets,'' in
  \emph{Advances in neural information processing systems}, 2014, pp.
  2672--2680.

\bibitem{Zhizheng13}
Z.~Wu, E.~S. Chng, and H.~Li, ``Conditional restricted boltzmann machine for
  voice conversion,'' in \emph{ChinaSIP}, 2013.

\bibitem{Nakashika2015}
T.~Nakashika, T.~Takiguchi, and Y.~Ariki, ``Voice conversion using
  speaker-dependent conditional restricted boltzmann machine,'' \emph{EURASIP
  Journal on Audio, Speech, and Music Processing}, vol. 2015, no.~1, p.~8,
  2015.

\bibitem{Nakashika2016}
T.~Nakashika, T.~Takiguchi, Y.~Minami, T.~Nakashika, T.~Takiguchi, and
  Y.~Minami, ``Non-parallel training in voice conversion using an adaptive
  restricted boltzmann machine,'' \emph{IEEE/ACM Trans. Audio, Speech and Lang.
  Proc.}, vol.~24, no.~11, pp. 2032--2045, Nov. 2016.

\bibitem{radford2015unsupervised}
A.~Radford, L.~Metz, and S.~Chintala, ``Unsupervised representation learning
  with deep convolutional generative adversarial networks,'' \emph{arXiv
  preprint arXiv:1511.06434}, 2015.

\bibitem{larsen2015autoencoding}
A.~B.~L. Larsen, S.~K. S{\o}nderby, H.~Larochelle, and O.~Winther,
  ``Autoencoding beyond pixels using a learned similarity metric,'' \emph{arXiv
  preprint arXiv:1512.09300}, 2015.

\bibitem{blaauw2016modeling}
M.~Blaauw and J.~Bonada, ``Modeling and transforming speech using variational
  autoencoders,'' \emph{Interspeech 2016}, pp. 1770--1774, 2016.

\bibitem{hsu2016voice}
C.-C. Hsu, H.-T. Hwang, Y.-C. Wu, Y.~Tsao, and H.-M. Wang, ``Voice conversion
  from non-parallel corpora using variational auto-encoder,'' in
  \emph{Asia-Pacific Signal and Information Processing Association Annual
  Summit and Conference (APSIPA)}.\hskip 1em plus 0.5em minus 0.4em\relax IEEE,
  2016, pp. 1--6.

\bibitem{van2016wavenet}
A.~van~den Oord, S.~Dieleman, H.~Zen, K.~Simonyan, O.~Vinyals, A.~Graves,
  N.~Kalchbrenner, A.~Senior, and K.~Kavukcuoglu, ``Wavenet: A generative model
  for raw audio,'' \emph{CoRR abs/1609.03499}, 2016.

\bibitem{mehri2016samplernn}
S.~Mehri, K.~Kumar, I.~Gulrajani, R.~Kumar, S.~Jain, J.~Sotelo, A.~Courville,
  and Y.~Bengio, ``Sample{RNN}: An unconditional end-to-end neural audio
  generation model,'' \emph{arXiv preprint arXiv:1612.07837}, 2016.

\bibitem{tan2016learning}
S.~Tan and K.~C. Sim, ``Learning utterance-level normalisation using
  variational autoencoders for robust automatic speech recognition,'' in
  \emph{Spoken Language Technology Workshop (SLT), 2016 IEEE}.\hskip 1em plus
  0.5em minus 0.4em\relax IEEE, 2016, pp. 43--49.

\bibitem{harwath2017learning}
D.~Harwath and J.~R. Glass, ``Learning word-like units from joint audio-visual
  analysis,'' \emph{arXiv preprint arXiv:1701.07481}, 2017.

\bibitem{garofolo1993darpa}
J.~S. Garofolo, L.~F. Lamel, W.~M. Fisher, J.~G. Fiscus, and D.~S. Pallett,
  ``{DARPA TIMIT acoustic-phonetic continous speech corpus CD-ROM. NIST speech
  disc 1-1.1},'' \emph{NASA STI/Recon technical report n}, vol.~93, 1993.

\bibitem{zue1990speech}
V.~Zue, S.~Seneff, and J.~Glass, ``{Speech database development at MIT: TIMIT
  and beyond},'' \emph{Speech Communication}, vol.~9, no.~4, pp. 351--356,
  1990.

\bibitem{kingma2014adam}
D.~Kingma and J.~Ba, ``Adam: A method for stochastic optimization,''
  \emph{arXiv preprint arXiv:1412.6980}, 2014.

\end{thebibliography}

\end{document}